\documentclass[10pt,twocolumn,letterpaper]{article}

 \usepackage{cvpr}              % To produce the CAMERA-READY version
\definecolor{cvprblue}{rgb}{0.21,0.49,0.74}
\usepackage[pagebackref,breaklinks,colorlinks,allcolors=cvprblue]{hyperref}

\def\paperID{00016} % *** Enter the Paper ID here
\def\confName{CVPR}
\def\confYear{2025}

\title{FoodTrack: Estimating Handheld Food Portions with
Egocentric Video}

\author{Ervin Wang\\
University of Waterloo\\
{\tt\small e92wang@uwaterloo.ca}
\and
Yuhao Chen \\
University of Waterloo \\
{\tt\small yuhao.chen1@uwaterloo.ca}
}

\begin{document}
\maketitle
\begin{abstract}

Accurately tracking food consumption is crucial for nutrition and health monitoring. Traditional approaches typically require specific camera angles, non-occluded images, or rely on gesture recognition to estimate intake, making assumptions about bite size rather than directly measuring food volume. We propose the FoodTrack framework for tracking and measuring the volume of hand-held food items using egocentric video which is robust to hand occlusions and flexible with varying camera and object poses. FoodTrack estimates food volume directly, without relying on intake gestures or fixed assumptions about bite size, offering a more accurate and adaptable solution for tracking food consumption. We achieve absolute percentage loss of approximately $7.01\%$ on a handheld food object, improving upon a previous approach that achieved a $16.40\%$ mean absolute percentage error in its best case, under less flexible conditions.

\end{abstract}    
\section{Introduction}
\label{sec:intro}

Accurately tracking food consumption is essential for nutrition and health monitoring. Traditional methods like self-reported food diaries or recall surveys often suffer from inaccuracies and biases \cite{ravelli2020traditional}. Automated approaches have used sensors \cite{farooq2019validation} and intake gestures \cite{tang2022new, wang2023eat} to infer nutritional intake, typically estimating the number of bites rather than the actual food volume. Wearable devices have also been used to monitor food intake through detected gestures \cite{fontana2014automatic}, relying on hand positioning or mouth contact to identify eating events \cite{wang2022eating}. Unlike gesture-based methods, 3D food reconstruction offers more precise volume estimation, but many existing algorithms require specific camera angles \cite{abdur2023comparative} or numerous images from multiple viewpoints \cite{makhsous2019novel}, making them impractical for real-world dietary monitoring where food is often manipulated by hand and subject to occlusions.

Our primary contribution is the development of an end-to-end pipeline for estimating the volume of hand-held food items from egocentric video. This method first captures video data through the Project Aria glasses \cite{engel2023project}, and enhances it through super-resolution. Segmentation and depth masks are then generated for each RGB video frame. From here, BundleSDF \cite{wen2023bundlesdf} is used to generate an estimated 3D mesh of the food object. To determine the true scale of the object, a single-frame absolute depth estimation model is used, and subsequently the volume of the generated mesh is scaled to compute the actual volume. We show an experimental result of $7.01\%$ absolute percentage error between the volume of a sandwich object reconstructed using our method and the measured volume of the sandwich, which is an improvement compared to a previous approach \cite{abdur2023comparative} that achieved a best-case $16.40\%$ mean absolute percentage error, under stricter data collection conditions.

\section{Framework and Method}
\label{sec:framework}

The framework minimizes user involvement for food volume estimation using Project Aria glasses to noninvasively capture data. It uses BundleSDF to generate 3D food meshes without requiring object or interaction priors, reducing the impact of hand occlusions. 

It was observed that without super-resolution, selecting and matching keypoints in food images is challenging due to blurring. However, accurate keypoint detection and matching are crucial for BundleSDF to effectively track the object's pose over time. Therefore, given a monocular RGB input video from a pinhole camera obtained by transforming data captured from the Aria glasses, our method begins with augmenting captured video data with super-resolution. Subsequently, a zero-shot segmentation method is employed to generate the initial segmentation mask, and a video object segmentation network is used to create segmentation masks for each frame. When using per-frame depth estimation there is considerable disparity between neighboring frames, which could negatively impact the accuracy of later object volume estimation. To address this issue, we opted to use a temporal depth estimation method to generate more consistent depth maps. The specific tools used were ResShift \cite{yue2024resshift} for super-resolution, Grounded SAM \cite{ren2024grounded} for zero-shot segmentation, Cutie \cite{cheng2024putting} for video object segmentation, and ChronoDepth \cite{shao2024learning} for depth estimation. See Fig.~\ref{fig3}, Fig.~\ref{fig4}, and Fig.~\ref{fig5} for example RGB, mask, and depth images. When using Grounded SAM, the specific prompt used was "food without background". This prompt was used in order to avoid wrapping paper/containers.

\begin{figure}[h]
    \centering
    \includegraphics[scale=0.05]{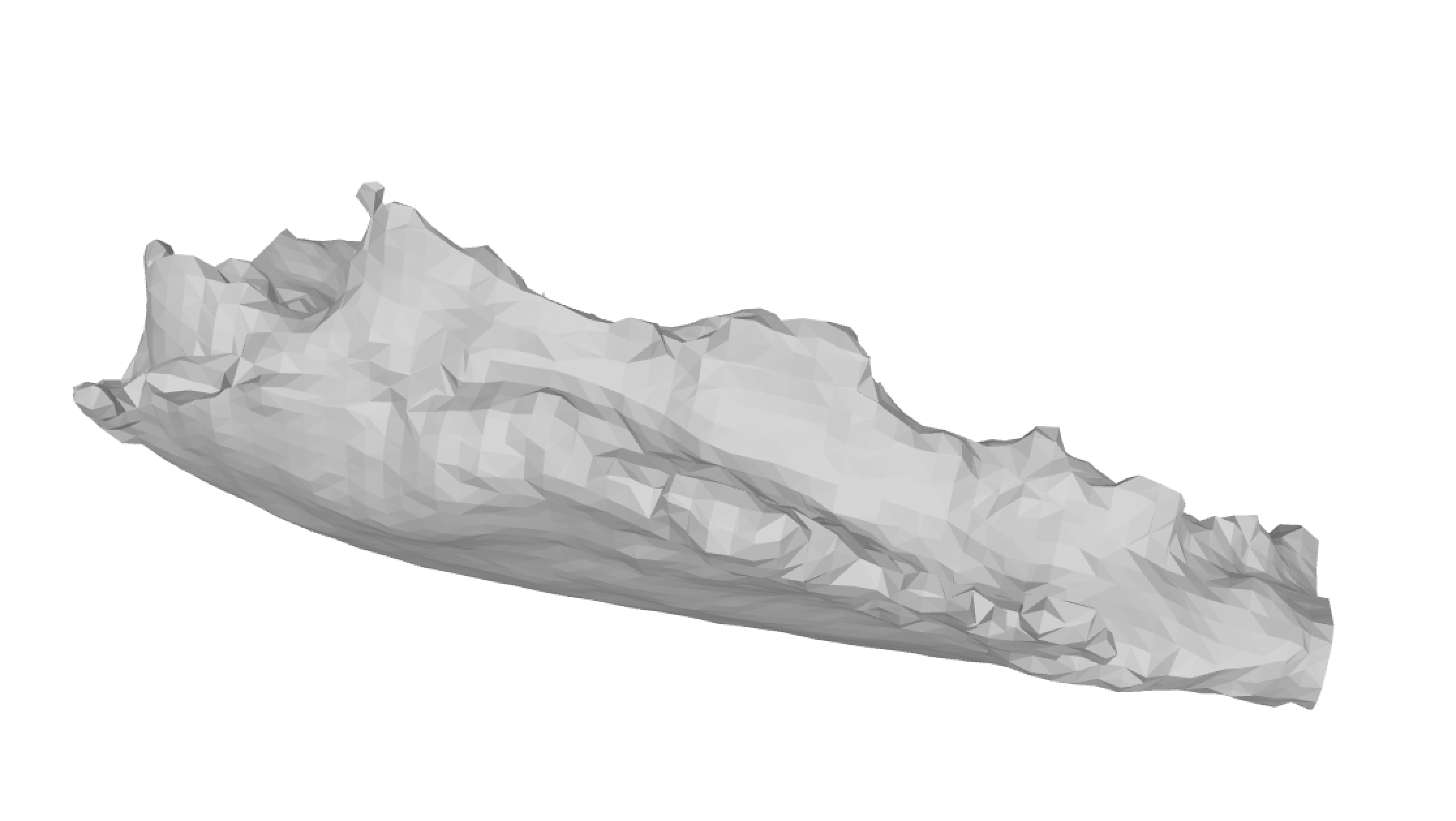}
    \includegraphics[scale=0.05]{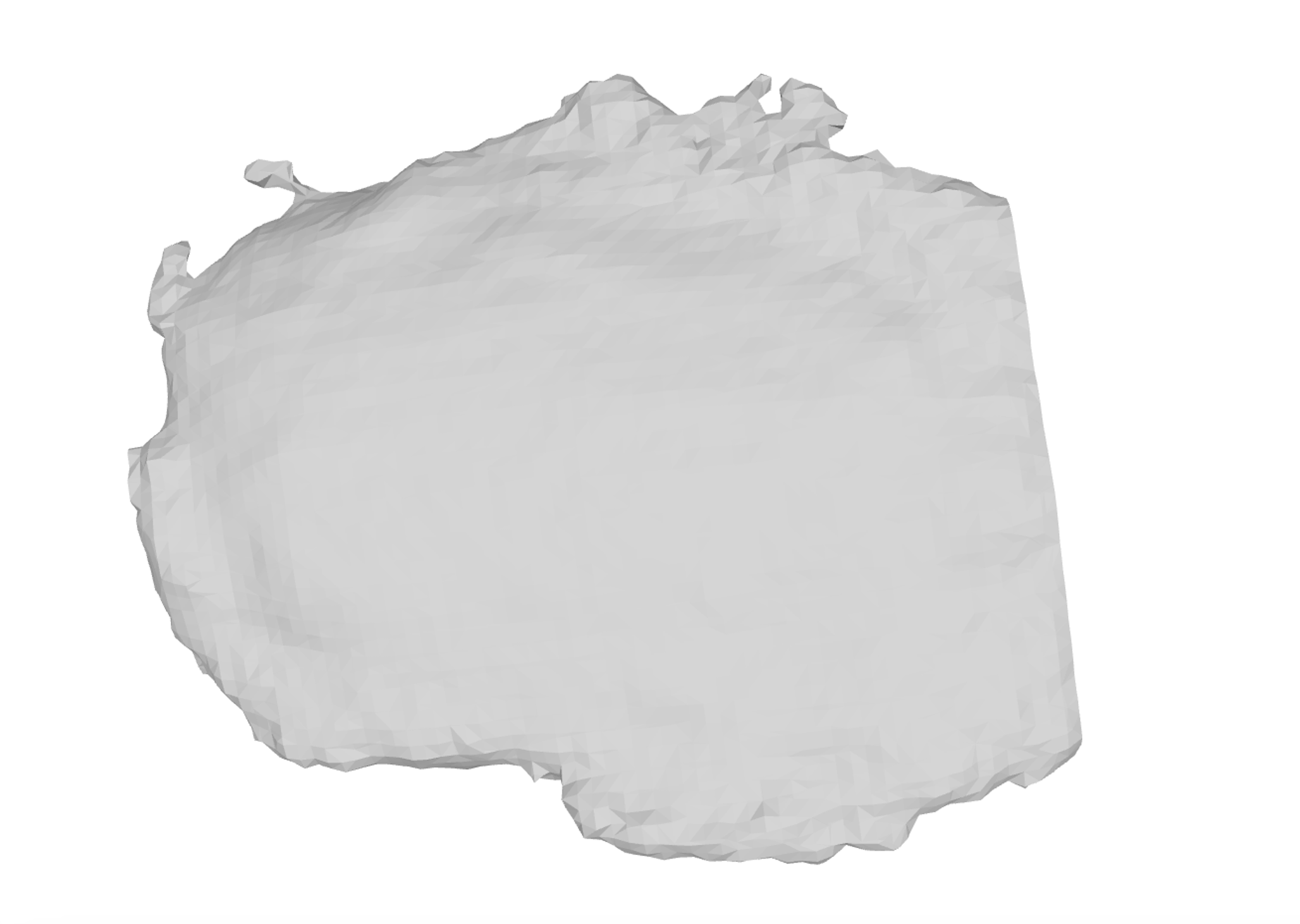}
    \caption{Reconstructed mesh without super-resolution}
    \label{fig1}
\end{figure} 
\begin{figure}[h]
    \centering
    \includegraphics[scale=0.1]{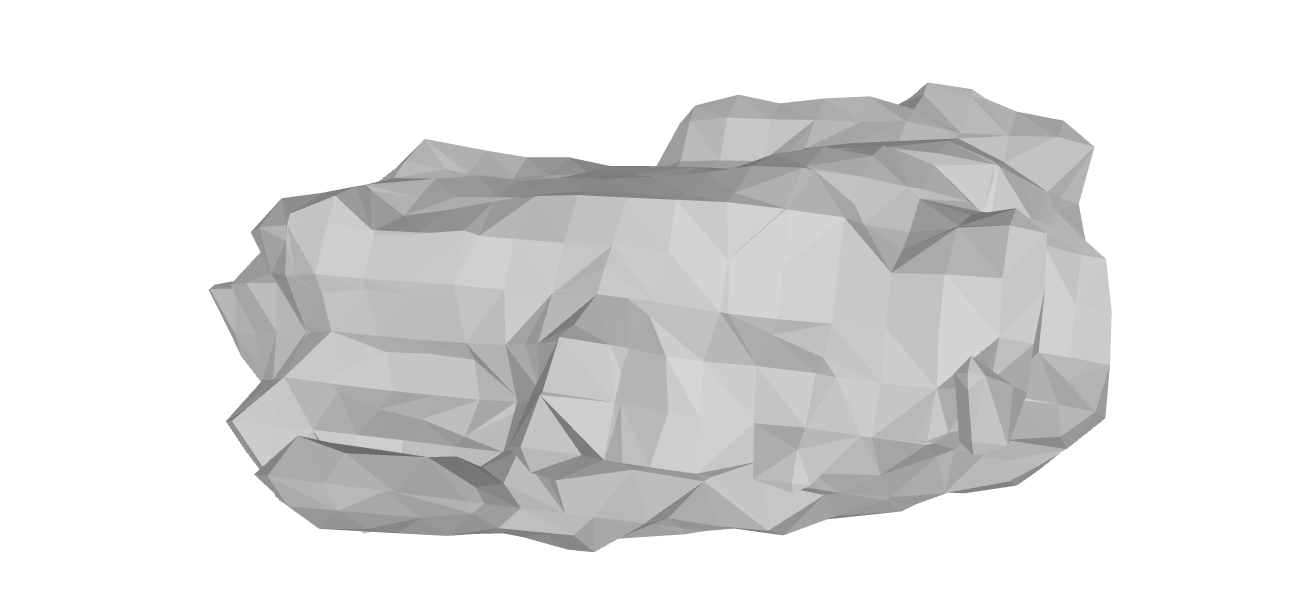}
    \includegraphics[scale=0.1]{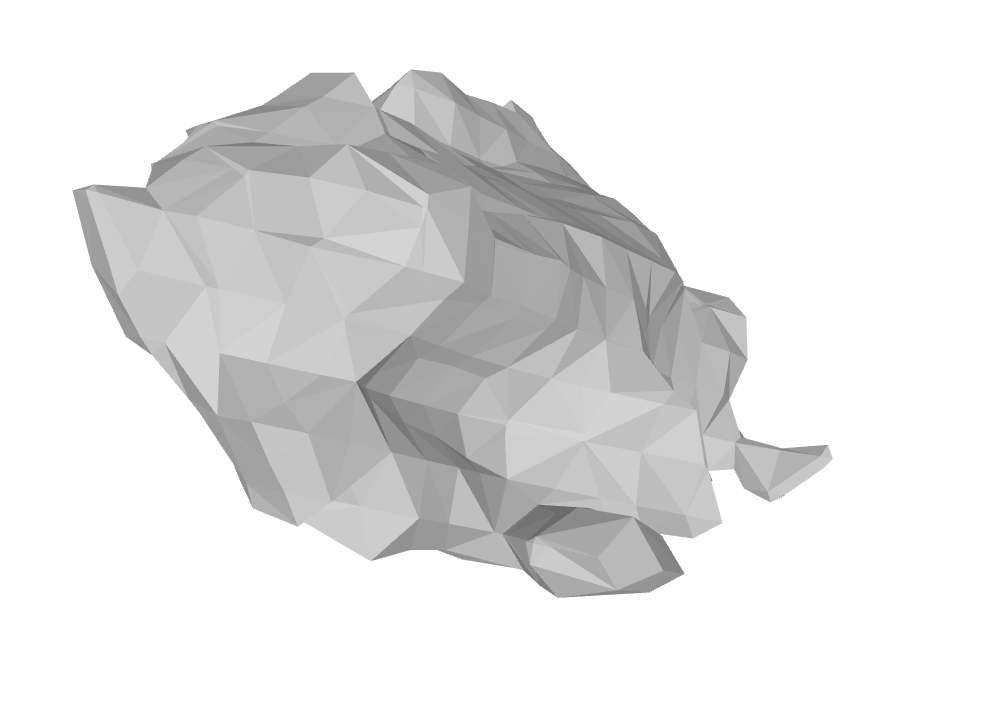}
    \caption{Reconstructed mesh with super-resolution}
    \label{fig2}
\end{figure} 

\begin{figure}[h]
    \centering
    \includegraphics[scale=0.08]{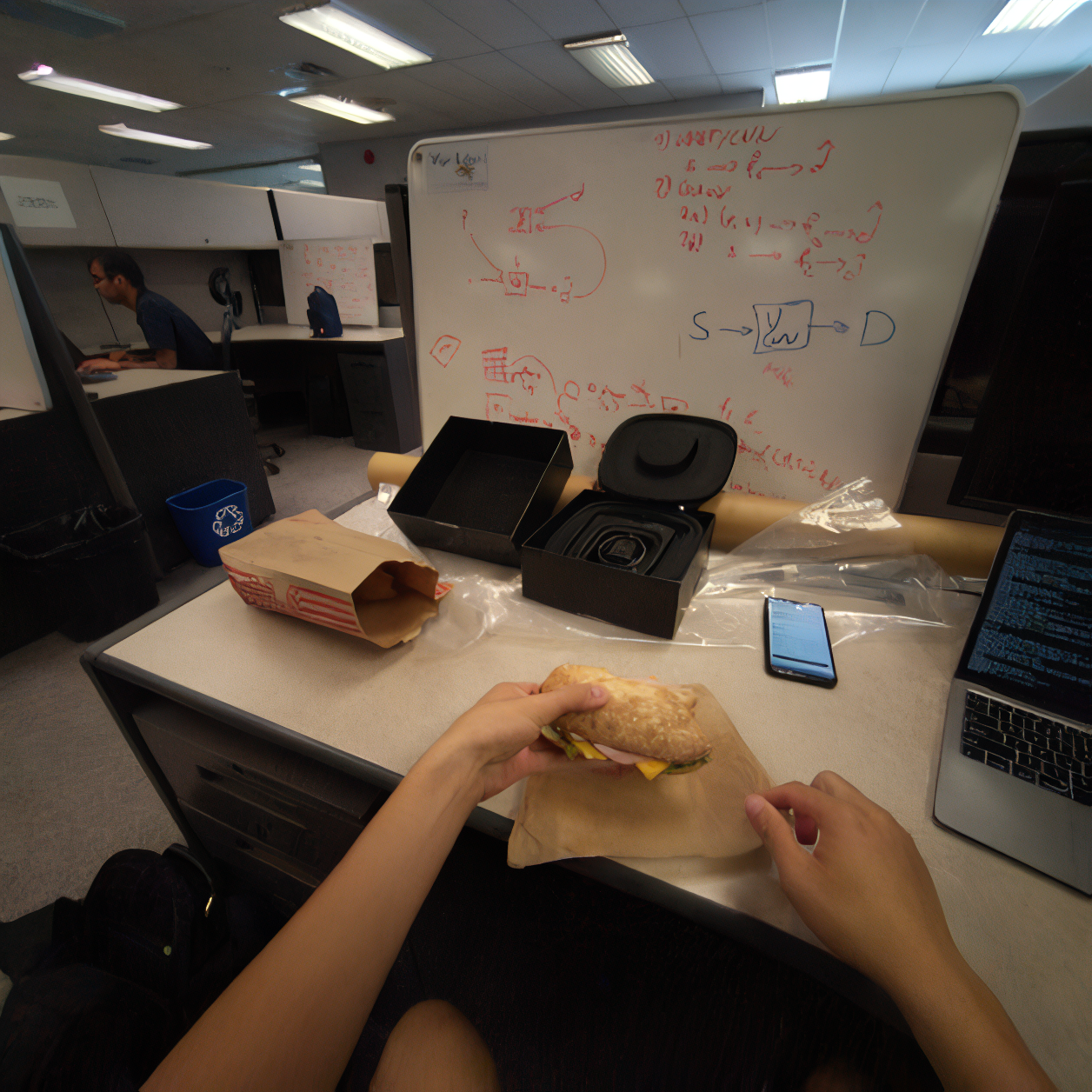}
    \caption{RGB image of an egocentric scene with two visible hands, and one hand is holding a sandwich }
    \label{fig3}
\end{figure} 
\begin{figure}[h]
    \centering
    \includegraphics[scale=0.08]{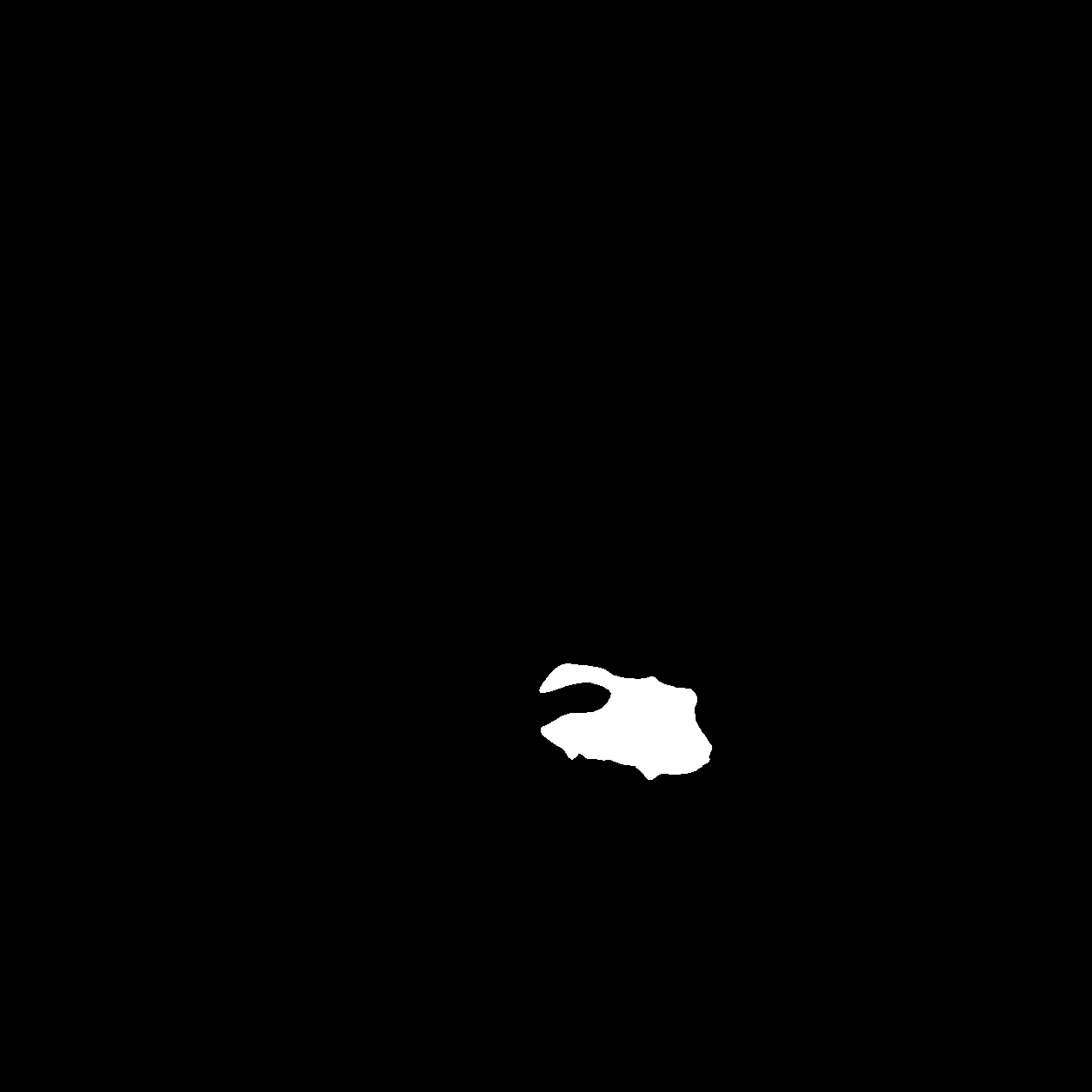}
    \caption{Binary mask of sandwich generated by Cutie from Fig.~\ref{fig3}}
    \label{fig4}
\end{figure}
\begin{figure}[h]
    \centering
    \includegraphics[scale=0.08]{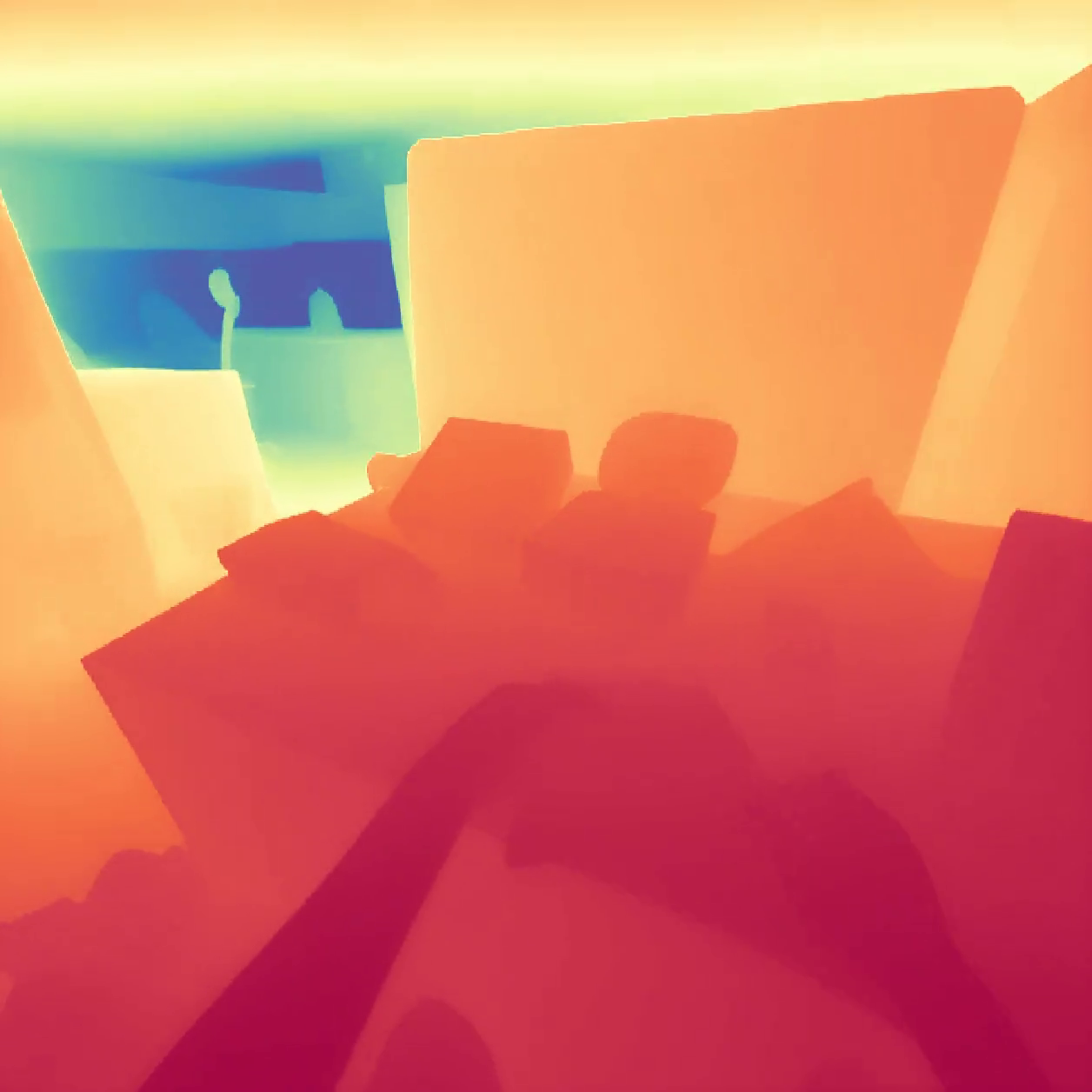}
    \caption{Depth image generated by ChronoDepth from Fig.~\ref{fig3}.}
    \label{fig5}
\end{figure}

\begin{figure*}[h]
    \centering
    \includegraphics[scale=0.52]{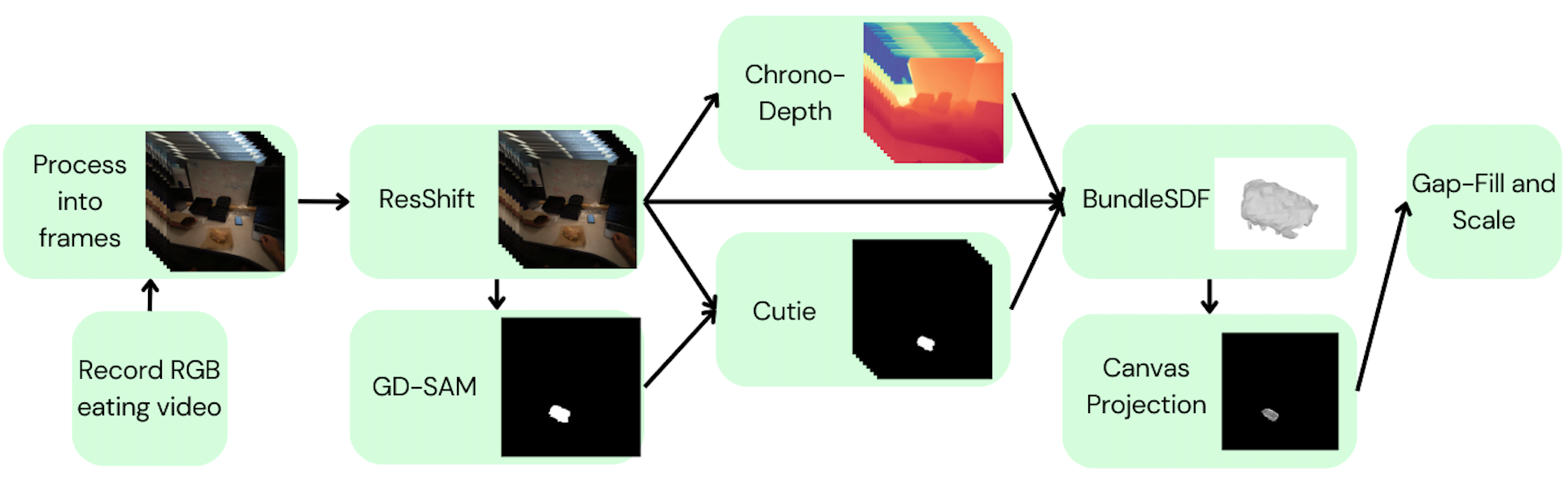}
    \caption{Processing pipeline for portion estimation.}
    \label{fig7}
\end{figure*}

After applying super-resolution, keypoint matching improves, but we still encounter a significant number of incorrect correspondences. To address this, we replace the matching algorithm in BundleSDF with LightGlue \cite{lindenberger2023lightglue} for greater accuracy. We then apply our modified version of BundleSDF to the processed data. Non-waterproof meshes are frequently generated by the BundleSDF algorithm. To ensure they possess a defined volume, we gap-fill them by using PyVista's \text{fill\_holes} method, which triangulates the holes to fill it with new triangular faces. The volume of the mesh is then computed using the \text{mesh.volume} method provided by Trimesh. The scale of the generated mesh by BundleSDF is incorrect, so we project it onto a plane such that the orientation aligns with the actual object and the $z$-axis values of the projected vertices preserve the relative depth information. Thus, we can scale the object based on the dimensions in the $x$ and $y$ directions. To get the projection, a view matrix and a camera projection matrix, $V_c, P_c$ respectively, are applied sequentially to vertices of the generated mesh $M$ to transform them onto a canvas, ensuring that the pose of the object on the canvas is the same as the pose in the RGB image. For simplicity in the following calculations, square images are used.

For each frame, the BundleSDF algorithm returns an estimated pose $p$ relative to the first frame, which is taken to be at origin pose. By inverting this pose, the view matrix $V_c$ relative to the object is found. That is, 

\begin{equation}
V_c = p^{-1}
\end{equation}

Given the camera intrinsic matrix \begin{equation}
K=[K_{ij}]_{(1\leq i,j \leq 4)}
\end{equation} and taking $n$ to be the near plane distance, and $f$ to be the far plane distance, the projection matrix $P_c$ is calculated as follows:

\renewcommand{\arraystretch}{1.5}
\begin{equation}
P_c = \begin{bmatrix}
\frac{2K_{00}}{W} & \frac{-2K_{01}}{W} & \frac{W - 2K_{02}}{W} & 0 \\
0 & \frac{-2K_{11}}{H} & \frac{H - 2K_{12}}{H} & 0 \\
0 & 0 & \frac{-f - n}{f - n} & \frac{-2 \cdot f \cdot n}{f - n} \\
0 & 0 & -1 & 0
\end{bmatrix}
\end{equation}

\begin{figure}[h]
    \centering
    \includegraphics[scale=1]{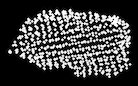}
    \includegraphics[scale=1]{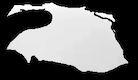}
    \caption{Example of projected and scaled mesh beside its respective mask.}
    \label{fig6}
\end{figure}

Taking $M_i$ as the $i$th mesh point, the transformation

\begin{equation}
M'_{i} = P_cV_cM_i
\end{equation}

is applied in order to project the mesh vertices onto a canvas. Call the entire projected mesh as $M'$. Since the video is from a pinhole camera, to enable direct estimation of the object's absolute size, we apply the ratio of an estimated focal length and object depth to the object size in pixel coordinates. This is achieved by first transforming the mesh to accurately reflect its size relative to the pixel coordinates in the RGB image. It is more straightforward to compute transformations from a normalized mesh to pixel coordinates so the mesh is normalized by considering the object’s relative size in the RGB image, and then transformed to an un-normalized pixel space where the $x$ and $y$ values of the mesh vertices correspond to their actual positions in the RGB image. See Fig.~\ref{fig6} for an example of a projected mesh next to its mask image. To normalize the mesh, the maximum and minimum coordinates of the mesh along the $x$ and $y$ axes in the transformed space are computed. The mesh is scaled to occupy the same percent width within an axis-aligned cube centered at the origin with side length 2, as the object in the image. Its distance from the original is also scaled by the same factor, effectively normalizing it based on its relative width in the RGB image. That is, defining the object pixel width as $w_{OP}$ and the image pixel width as $w_{IP}$, and $M^{(N)}$ as the mesh scaled to the normalization cube, scale the mesh following the formula below:

\begin{equation}
M^{(N)}_i = (M'_i - \text{Centroid}(M')) \cdot S_1 + \text{Centroid}(M') \cdot \frac{w_{OP}}{w_{IP}}
\end{equation}

To transform the normalized mesh to an un-normalized pixel space, taking $M^{(RGB)}$ to be the mesh with vertices projected to their actual positions in the RGB image and letting $M^{(N)}_{i,x},M^{(N)}_{i,y},M^{(N)}_{i,z}$ be the $x,y,z$ coordinates of the $i$th vector in $M^{(N)}$, we compute

\begin{equation}
M^{(RGB)}_{i,x} = \frac{M^{(N)}_{i,x}+1}{2}\cdot L
\end{equation}

\begin{equation}
\text{For } w \in \{y,z\}, M^{(RGB)}_{i,w} = (1-\frac{M^{(N)}_{i,w}+1}{2}) \cdot L
\end{equation}

Once the mesh is in pixel space, we calculate the size ratio between meters and pixels to properly scale the volume of the mesh. The Depth-Pro \cite{bochkovskii2024depth} model is applied on the last image used to construct the mesh, to estimate the depth $D$ in meters of the object in the RGB image at the centroid of its 2D mask, as well as the camera’s focal length $f_x$. With this information, we compute the size ratio $R$ to be

\begin{equation}
R = \frac{D}{f_x}
\end{equation}

From here, the volume of $M^{(RGB)}$ is calculated using the inbuilt trimesh method, and multiplied by $(R)^3$ to get the estimated food volume. Fig.~\ref{fig7}. shows our processing pipeline to get the final volumes.

\section{Data Collection}

A 15-second video of a rotating sandwich was recorded using Project Aria glasses. To measure the sandwich’s volume, the water displacement method was used, with the sandwich wrapped in plastic to prevent water absorption. The displaced water volume provided the measurement.
\section{Preliminary Results}

\iffalse
\begin{figure}[h]
    \centering
    \includegraphics[scale=0.1]{images/Screenshot 2024-11-07 at 3.20.01 AM.png}
    \includegraphics[scale=0.1]{images/Screenshot 2024-11-07 at 3.20.14 AM.png}
    \includegraphics[scale=0.1]{images/Screenshot 2024-11-07 at 3.20.18 AM.png}
    \caption{Three viewpoints of estimated sandwich model.}
    \label{fig8}
\end{figure}
\fi

Empirically, the sandwich in Fig.~\ref{fig3} was measured to have a volume of $371 \pm 1$ mL, and the volume of the sandwich was estimated to be approximately 345 mL using the proposed method. The absolute percentage error $A$ is computed as

\begin{equation}
A = \frac{|345-371|}{|371|} \times 100\% \approx 7.01\%
\end{equation}

Although the analysis is based on a single example, the results suggest that the method shows promise and warrants further investigation given its potential for improved performance, compared to a previous approach \cite{abdur2023comparative} that achieved a $16.40\%$ mean absolute percentage error in its best case with more restrictive data gathering.
\section{Conclusion}

The proposed framework for tracking food volume from egocentric video offers significant improvements over traditional methods. However, challenges remain, particularly with BundleSDF, which struggles with accurate 3D reconstructions, especially for objects with rotationally invariant silhouettes.

Future work will focus on enhancing 3D reconstruction for complex objects and improving volume estimation for individual bites. The handheld nature of food will be leveraged to track its position relative to the hand, and pretrained models with food-related knowledge will be explored to increase accuracy.

This framework provides a more reliable and flexible approach to dietary monitoring, with strong potential for health applications.
{
    \small
    \bibliographystyle{ieeenat_fullname}
    \bibliography{main}
}

\end{document}